\documentclass[journal]{IEEEtran}

\usepackage{graphicx}
\usepackage{overpic}
\usepackage{xcolor}
\usepackage{bm}
\usepackage{amsmath}
\usepackage{amssymb}
\usepackage{bold-extra}
\usepackage{svg}
\usepackage[hyphens]{url}
\usepackage[backend=biber,style=ieee,natbib=true]{biblatex} 
\IEEEoverridecommandlockouts

\usepackage{tikz}
\usepackage{textcomp}
\usepackage{hyperref}
\usepackage{lipsum}

\newcommand\copyrighttext{%
  \footnotesize \textcopyright 2022 IEEE. Personal use of this material is permitted.
  Permission from IEEE must be obtained for all other uses, in any current or future
  media, including reprinting/republishing this material for advertising or promotional
  purposes, creating new collective works, for resale or redistribution to servers or
  lists, or reuse of any copyrighted component of this work in other works.}
\newcommand\copyrightnotice{%
\begin{tikzpicture}[remember picture,overlay]
\node[anchor=south,yshift=1pt] at (current page.south) {\fbox{\parbox{\dimexpr\textwidth-\fboxsep-\fboxrule\relax}{\copyrighttext}}};
\end{tikzpicture}%
}

\newcommand\rev[1]{\textcolor{black}{#1}}

\begin{document}
\title{{\sc Olivaw}: Mastering Othello without \rev{Human Knowledge, nor a Fortune}}

\author{Antonio Norelli, Alessandro Panconesi \\ 
Dept. of Computer Science, Università La Sapienza, Rome, Italy \thanks{Corresponding authour: Antonio Norelli (email: norelli@di.uniroma1.it).}}

\maketitle
\copyrightnotice

\begin{abstract}
We introduce {\bfseries\scshape Olivaw}, an AI Othello player adopting the design principles of the famous AlphaGo \rev{programs}. The main motivation behind {\bfseries\scshape Olivaw} was to attain exceptional competence in a non-trivial board game at a tiny fraction of the cost of its illustrious predecessors.
In this paper, we 
show how the AlphaGo Zero's paradigm can be successfully applied to the popular game of Othello using only commodity hardware and free cloud services.
While being simpler than Chess or Go, Othello maintains a considerable search space and difficulty in evaluating board positions.
To achieve this result, {\bfseries\scshape Olivaw} implements some improvements inspired by recent works to accelerate the standard AlphaGo Zero learning process.
The main modification implies doubling the positions collected per game during the training phase, by including also positions not played but largely explored by the agent.
We tested the strength of {\bfseries\scshape Olivaw} in three different ways: 
by pitting it against Edax, considered by many the strongest open-source Othello engine,
by playing anonymous games on the web platform OthelloQuest, and finally in two in-person matches against top-notch human players: a national champion and a former world champion.

\end{abstract}


\IEEEpeerreviewmaketitle

\section{Introduction}

\IEEEPARstart{O}{nly} a year after AlphaGo's landmark victory against Go master Lee Sedol another sensational development took place.  
An improved version of  AlphaGo called AlphaGo Zero asserted itself as the strongest Go player in the history of the game  \cite{Silver2017MasteringKnowledge}. The remarkable feature of  AlphaGo Zero was that, unlike its predecessor and unlike all previous game software,  it learned to master the game entirely by itself, without any human knowledge. As subsequent follow-up work quickly showed, AlphaGo's paradigm-- an interesting blend of deep and reinforcement learning-- seems to be general and flexible enough to adapt to a wide array of games
\cite{Silver2018ASelf-play.}, \cite{muzero-schrittwieser2019mastering}.

These extraordinary successes came at a price however, and quite literally so. The amount of computational and financial resources that were required was so huge as to be out of reach for most academic and non-academic institutions. Not coincidentally these well-endowed projects and their follow-ups took place within giant multinational corporations of the IT sector \cite{tian2019elf,leeminigo}.  These companies deployed \rev{GPUs} by the thousands and hundreds of \rev{TPUs}. A recent study looked at the number of 
petaflops per day  that were required to train AlphaGo Zero and other recent well-known results in AI \cite{amodei2018ai}. The paper shows an exponential growth with a 3.4-month doubling period. This is clearly unsustainable for most academic labs and departments and even the greatest majority of companies. Another aspect of the same problem is the amount of training needed. 
AlphaGo Zero required 4.9 million games played during self-play. And in order
to attain the level of grandmaster for games like Starcraft II and Dota 2 the training required 200 years and more than $10,000$ years of gameplay, respectively \cite{vinyals2019grandmaster}, \cite{pachockiopenai}.

Thus one of the major problems to emerge in the wake of these breakthroughs is whether comparable results can be attained at a much lower computational and financial cost and with just commodity hardware. In this paper we take a small step in this direction, by showing that AlphaGo Zero's successful paradigm can be replicated for the game of Othello (also called Reversi). While being much simpler than either Chess or Go, this game is still rather sophisticated and has a considerable strategic depth.  The game enjoys a long history and a rich tradition. Every year an exciting world championship takes place in which accomplished players from all over the world vie for the world title.  

Our Othello engine is called {\sc Olivaw}, a homage to the famous robot character invented by Isaac Asimov.
We tested the strength of {\sc Olivaw} in three different ways. 
In one instance, we pitted {\sc Olivaw} against Edax, one of the strongest Othello engines. Perhaps the most interesting aspect of this set of matches was that {\sc Olivaw} managed to beat several times an opponent that explores tens of millions of positions in the game tree in the course of a single game. In contrast, {\sc Olivaw}'s search of the game tree was limited to a couple of thousand positions.

We also tested {\sc Olivaw} against (presumably) human players of varying strength on the web platform OthelloQuest. But the most exciting challenge was a series of matches against top-notch human players: a national champion and a former world champion.

The final outcome shows that in a relatively short training time {\sc Olivaw} reached the level of the best human players in the world. Crucially, this has been achieved by using very limited resources at very low overall cost: commodity hardware and free, and thus very limited, cloud services.

\section{Related work}

The success of AlphaGo naturally stimulated several follow-ups.
One of the main questions was to determine the level of generality of the approach. A series of papers showed this level to be great indeed. 
One after the other a list of difficult games fell pray of the RL-with-oracle-advice approach.
\citet{Silver2018ASelf-play.} extended it to Chess and Shogi. 

Recently \citet{muzero-schrittwieser2019mastering} added ATARI games to the list.  Our work continues this line of research by adding $8 \times 8$ Othello to the list,  paying special attention to the cost issue. 
Indeed, it is not clear a priori whether the approach scales down in terms of resources.  Although cheaper in some ways, the agents in  \cite{Silver2018ASelf-play.}, \cite{muzero-schrittwieser2019mastering},
\cite{tian2019elf}, and \cite{leeminigo}
still use thousands of GPU's or hundreds of TPU's to master board games.
The recent KataGo \cite{katago-wu2019accelerating} reaches the level \rev{of play} of ELF using 1/50 of the computation and implements several techniques to accelerate the learning. However, these include a set of targets crafted by humans which are very game-specific\footnote{For instance, a ladder indicator, where ladders are Go-specific tactics}
thereby reducing the generality of the approach and reintroducing human knowledge in a relevant way.

Successful low-cost reproductions of AlphaGo Zero came out in recent years, but only for very simple games like Connect-4 \cite{connect4MediumLessonsAlphaZero} or $6 \times 6$ Othello \cite{chang2018big}, for which perfect strategies are known.

Other works focused on the hyperparameters involved in AlphaGo Zero, looking for a faster and cheaper training process. \citet{wang2019hyper} and \citet{connect4MediumLessonsAlphaZero} make several experiments in this direction, while \citet{wu2020population-based-accelerating} investigates the possibility of tuning hyperparameters within a single run, using a population-based approach on Go. 
We followed some insights provided by these works as reported in section \ref{sec-training_process}.

Concerning the state-of-the-art of $8 \times 8$ Othello engines,
algorithms became superhuman before the deep learning era, a fact heralded by the defeat of the world champion Takeshi Murakami at the hands of Logistello \cite{buro1995logistello}. Today the strongest and most popular programs used by top Othello players for training and game analysis are Saio \cite{RomanoBenedetto2009SAIO:Dellothello} and the open source Zebra \cite{zebra} and Edax \cite{Edax}. 
As in chess before AlphaZero \cite{Kasparov2018ChessReasoning.},
in order to reach the level of world-class players, Othello engines rely on a highly optimized minimax search \cite{neumann1928theorie} and employ handcrafted evaluation functions based on knowledge of the game accumulated by human beings\footnote{
For instance, patterns on the edge of the corner of the board, which are known to be of great importance in Othello.} \cite{zebra}. 
Another key feature used by these more traditional engines are huge catalogs of opening sequences, distilled and stored in the course of decades by human players and, more recently, by software as well.
Finally, typically these engines play the perfect game by sheer computational brute force by expanding the entire game tree for a large portion of the game, typically starting 20-30 moves before the end.
To summarize, unlike {\sc Olivaw}, traditional Othello engines make crucial use of knowledge of the game accumulated by humans over many years and of a massive brute force tree exploration. These are two important limitations {\sc Olivaw} is attempting to overcome.

To the best of our knowledge, the only approach similar to ours in spirit is a paper by
\citet{liskowski2018learning} which presents an engine obtained by training a convolutional neural network (CNN) with a database of expert moves. The engine however was only able to defeat Edax 2-ply (tree search limited to two moves ahead), a level of Edax that is much weaker than top Othello human players.

\section{Othello}

\begin{figure}[h]
\begin{overpic}
		[trim=0cm 0cm 0cm 0cm,clip,width=1.\linewidth]{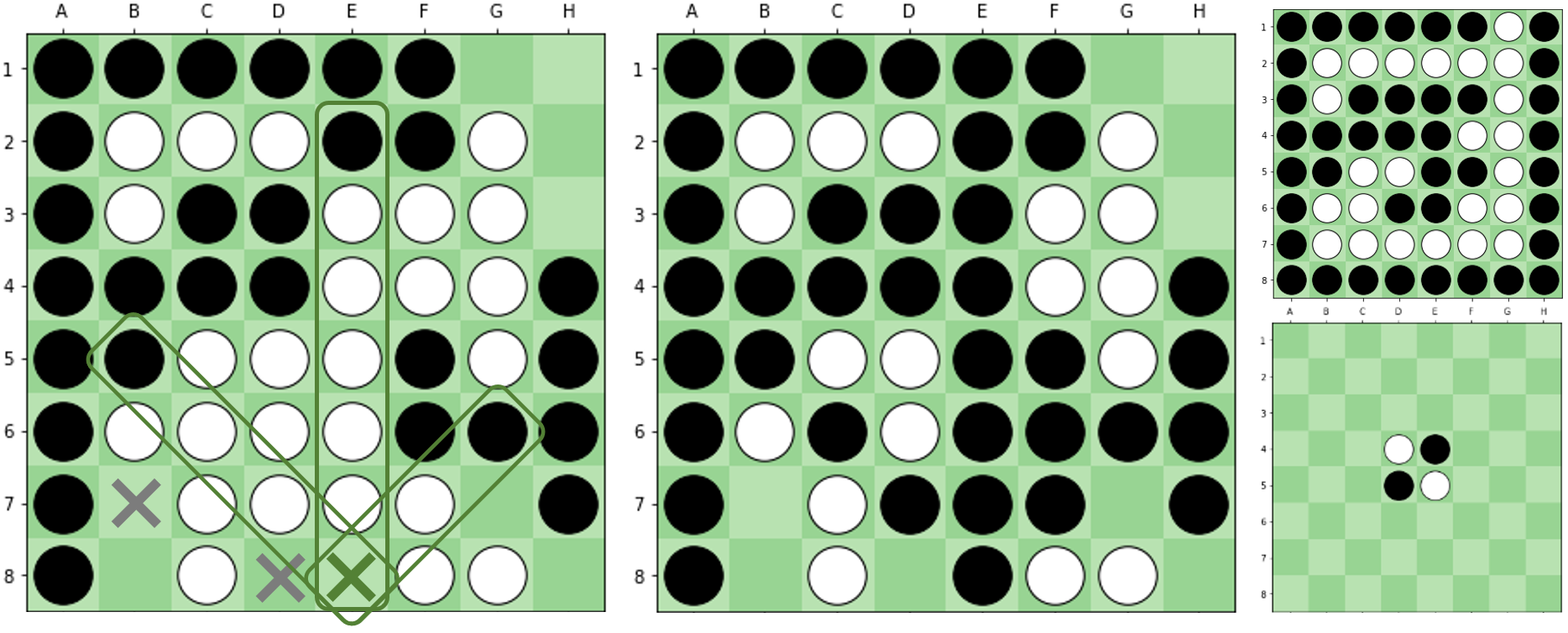}
		\put(18.7,-2.05){\scriptsize (a)  }
		\put(12,41){\scriptsize Turn of Black  }
		\put(51.9,41){\scriptsize Turn of White  }
		\put(88.8,41){\scriptsize (c)  }
		\put(58.8,-2.05){\scriptsize (b)  }
		\put(88.8,-2.05){\scriptsize (d)  }
\end{overpic}
\centering
\caption{\textbf{Rules of Othello.} Othello is a turn-based game where the black and white player try to overcome each other in the final domination of an 8x8 board. \textbf{a.} Players move alternately by placing a new disk in an empty square in order to bracket one or more opponent’s disks between the played disk and another of its own color already on the board. It is possible to capture disks horizontally, vertically, and diagonally. Disks can be captured in one or more directions in a single move, with capture always occurring in a straight line. Only moves capturing at least one disk are allowed. In the absence of moves the player must skip the turn. It is not possible to pass the turn if there is at least one valid move. \textbf{b.} The imprisoned disks change color and become owned by the player who moved. \textbf{c.} When none of the players can move, for instance when the board is full, the player with more disks on the board wins. Here black wins 40-24. \textbf{d.} A game of Othello begins with 4 disks placed in the center of the board in the shape of an X. Black moves first.
}
\label{fig:rules}
\end{figure}
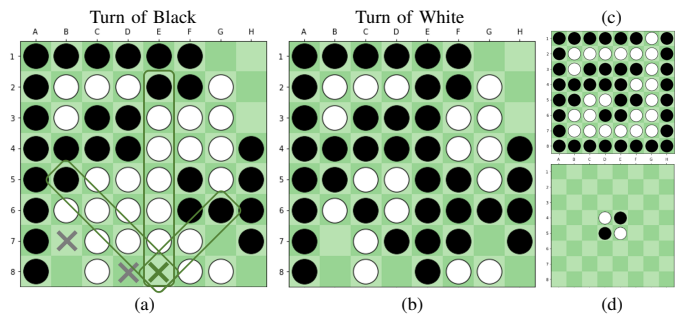

Othello is a popular board game. 
Its simple rules are explained in Figure \ref{fig:rules}.
A typical game lasts for some 60 moves, with an average branching factor of 10. Like Go and Chess it is a perfect information game. Although simpler than these two, it has considerable strategic depth. Unlike English draughts, there is no known perfect strategy that can be played by computer \cite{mullins2007checkers}. 

Othello is played across the globe. There are professional players competing in official tournaments organized by world and national federations. The Othello world championship takes place every year.

The best software beat humans systematically but, as discussed, they rely on brute force for most of the game. During the initial stages of the game, they access huge databases of openings, which are known to be very important in Othello. An opening lasts between 10 and 20 moves. Furthermore, some 20-30 moves before the end they play the perfect game by exploring the game tree in its entirety. In the middle game too, these programs explore hundreds of millions of positions of the game tree per move, a clear indication of brute force at play. 
In contrast, as we shall see, {\sc Olivaw} explores only a few hundreds positions and does not use any database of openings. 

\section{{\scshape Olivaw}: the algorithm}

The design of {\sc Olivaw}, our Othello engine, follows closely that of AlphaGo Zero \cite{Silver2017MasteringKnowledge}. The main difference consists of a somewhat different and cheaper training process. The network architecture, while mimicking that of AlphaGo Zero, was scaled down.
Before discussing {\sc Olivaw} in detail, it is useful to describe its basic design.

\subsection{The basic design}

Like the AlphaGo \rev{programs},  {\sc Olivaw} uses reinforcement learning to build an ``oracle'',
in the form of a deep network $f_\theta$ ($\theta$ denotes the weights of the neural network).  Given as input an Othello game state $s$, $f_\theta$ outputs a pair: $f_\theta(s) = (\bm{p}, v)$. The vector
$\bm{p}$ is a probability distribution over the possible moves from $s$. Intuitively, the higher the probability the better the move. The value $v$ is the oracle's assessment of how good state $s$ is, ranging from $+1$ (sure victory) to $-1$ (certain defeat). 

The oracle is used to guide an exploration of the 
\rev{``possible near futures''} by a Monte Carlo Tree Search (MCTS) \cite{mcts-browne2012survey}. To pick the next move, the game tree rooted at $s$ is explored. Roughly speaking, in this exploration, the moves that $f_\theta$ considers good are explored first (so that the  \rev{actual branching factor is limited}) and the total number of nodes explored is limited (in the few hundreds during training and set to one thousand when playing against humans).
The goal of this exploration phase is to produce a better estimate $(\bm{\pi}, q)$ of state $s$.
When this is done, the best move according to $\bm{\pi}$ is played to reach a new state $s'$, and the process is repeated.

What is noteworthy about this process is that while by itself $f_\theta$ is a rather weak player, using it in combination with MCTS gives rise to a very strong one, i.e. the estimates $(\bm{\pi}, q)$ are more reliable than $(\bm{p}, v)$.
Let us call $A(f)$ the MCTS playing agent using $f$ as oracle.

The crux of the approach is to generate a sequence of oracles
$f_0, f_1, f_2, \ldots, f_t$ each better than the predecessors. This is done by generating a sequence of training sets $S_1, S_2, \ldots, S_t$ each better than the previous one. Training set $S_i$ is used to train $f_i$. The process is initialized with a deep network $f_0$ with random weights.

The generic step in this sequence of improvements is as follows. Let $f_\theta$ be the current oracle. 
During the so-called \emph{self-play phase}, $A(f_\theta)$ plays a batch of games against itself.  During each game a set of states $S$ will be explored. For each $s \in S$ an updated (and hopefully better) assessment $(\bm{\pi}_s, q_s)$ will be computed.
The set $T$ of pairs $\{s, (\bm{\pi}_s, q_s)\}$ for $s \in S$ will be added to the training set. The intuition is that this way we can create a virtuous circle. As the assessments $(\bm{\pi}_s, q_s)$ become more and more accurate the training set becomes better and better. And, as the training set improves the assessment becomes more accurate.

We remark \rev{that} the main difference between {\sc Olivaw} and AlphaGo Zero resides in how this training set is constructed. Instead of 
$\{s, (\bm{\pi}_s, q_s)\}$, AlphaGo Zero only considers pairs 
$\{s, (\bm{\pi}_s, z_s)\}$ where $s$ is actually played during the game, and 
$z_s$ is the outcome at the end of the game. Thus, $z_s \in \{-1, 0, +1\}$. In contrast, besides this type of pairs, {\sc Olivaw} also adds to the training set pairs $\{s, (\bm{\pi}_s, q_s)\}$ for which $s$ has been explored ``a lot''. In this way, we collect a larger training set for the same number of simulated games.
This is crucial since the cost of the \emph{self-play} phase is the main contributor to the overall cost. This design choice is discussed in detail in section \ref{sec-comparison}.

Once the new training set is obtained we switch to the \emph{training phase}. The current neural network $f_\theta$ is further trained with the updated training set.
At the end of this training we have a new configuration $f_\theta'$. We want $(\bm p, v) = f_\theta'(s)$ to be close to $($\boldmath$\pi\,$\unboldmath$, \omega)$, where $\omega$ can be $z \;\mathrm{or}\; q$. So we minimize the training loss:

\begin{equation}
    L(\bm{\pi}, \omega, \bm{p}, v) = (\omega - v)^2 - \bm{\pi}^T\log{\bm{p}} + c||\theta||^2
\end{equation}

As in AlphaGo Zero, we combine evenly the squared error on the value and the cross-entropy on the move probabilities, while the last term penalizes large weights in the neural network ($c=10^{-4}$). This L2 regularization is used to prevent overfitting over the many training phases.

In the final \emph{evaluation phase}, we verify whether the new $f_\theta'$ is stronger than the old $f_\theta$. To do so, we pit $A(f_\theta')$ against $A(f_\theta)$.
If the former wins significantly more often than the latter it becomes the new oracle. Otherwise we go through the training phase again to produce a new challenger.
The whole process is then repeated. And so on, so forth.
For a more detailed description of the reinforcement learning algorithm we refer to \citet{Silver2017MasteringKnowledge}.

\begin{figure*}[t]

\includegraphics[width=0.32\textwidth]{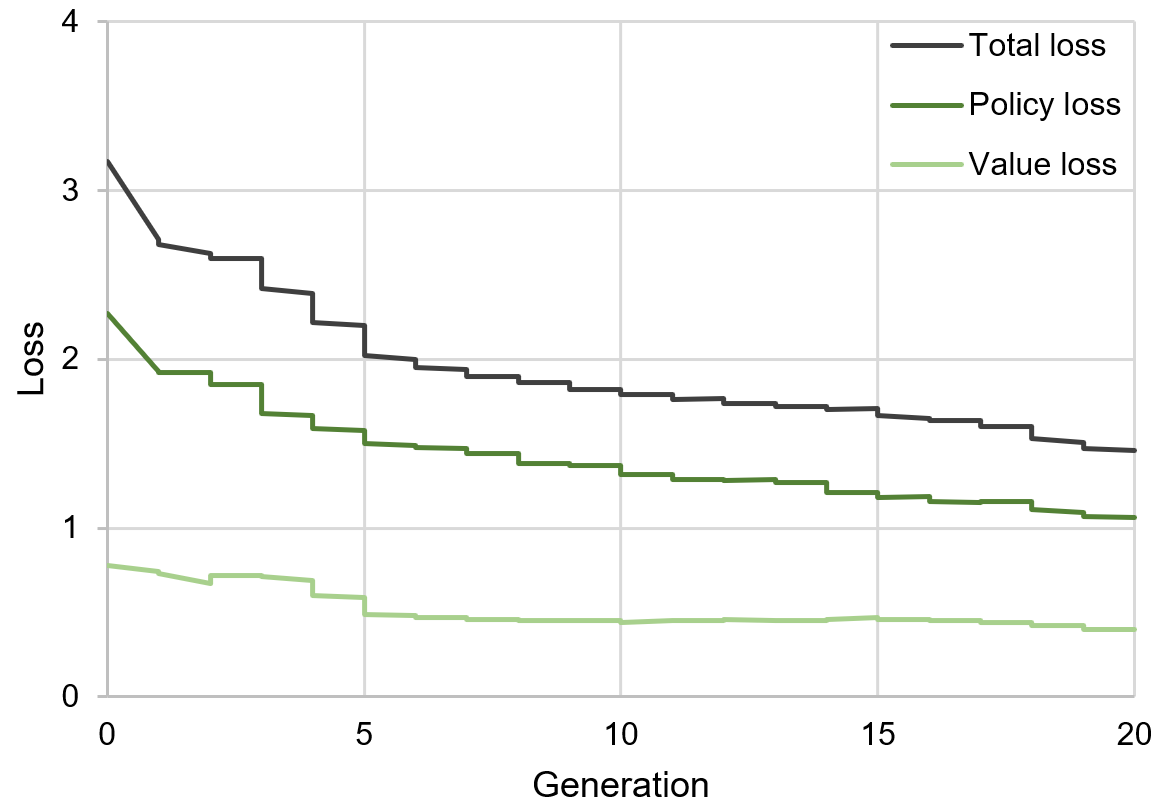}
\includegraphics[width=0.32\textwidth]{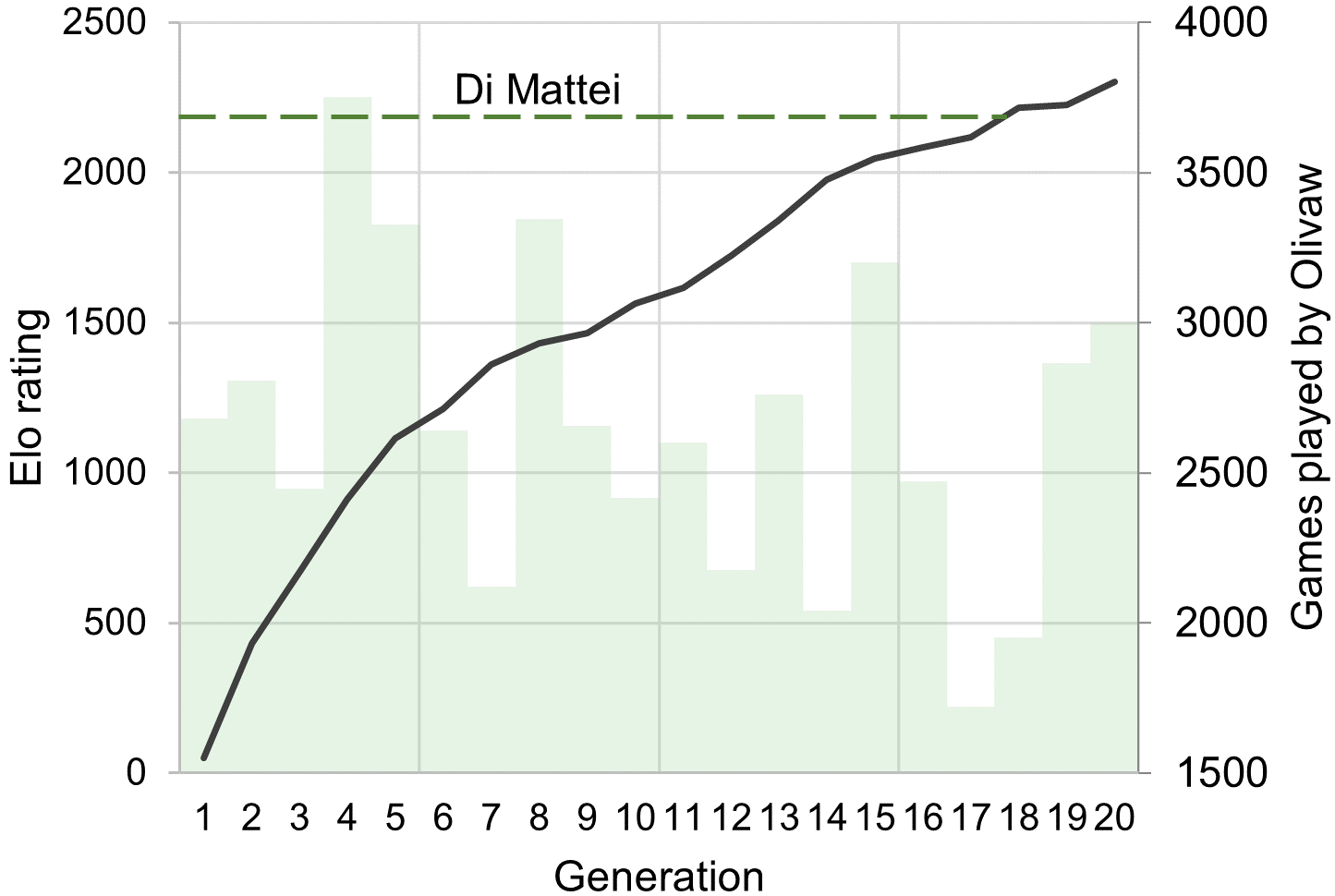}
\includegraphics[width=0.32\textwidth]{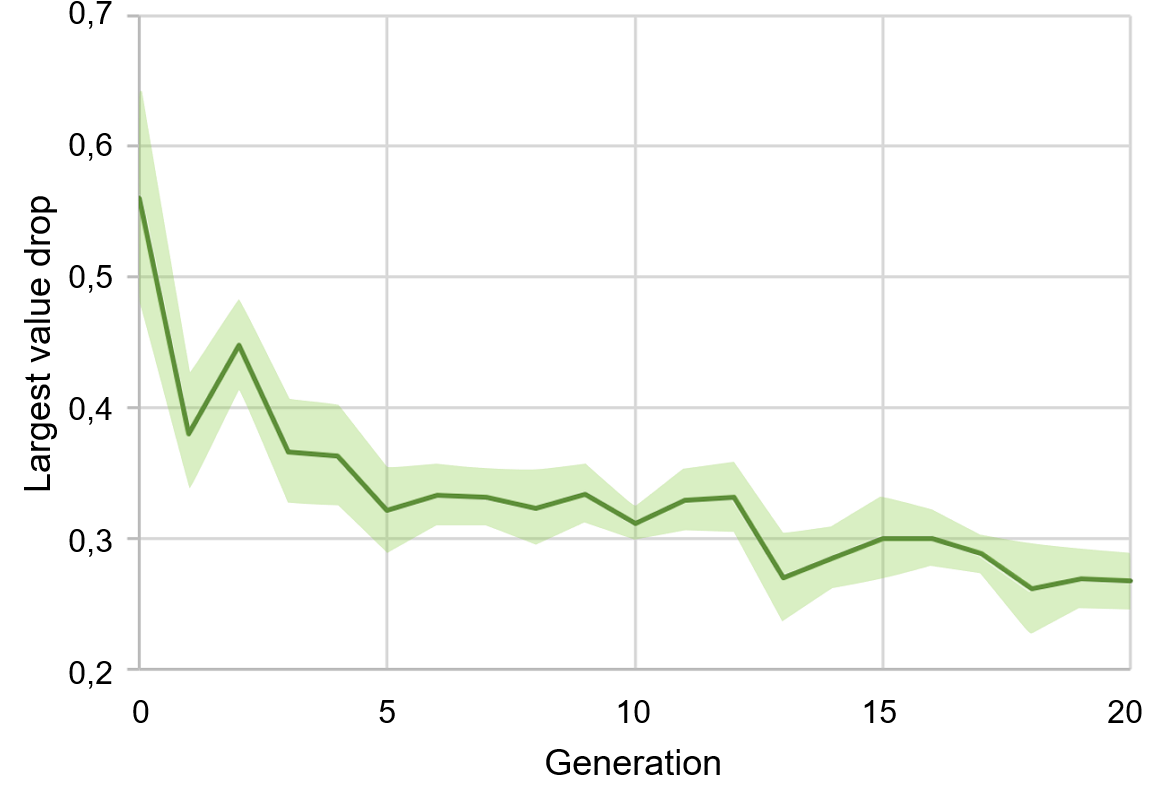}
\centering
\begin{overpic}
		[trim=10cm 34cm 2cm 0cm,clip,width=1.0\linewidth]{images/crucial_moves.png}
		\put(17.65,0){\scriptsize (a)  }
		\put(49.3,0){\scriptsize \rev{(b)}  }
		\put(83.6,0){\scriptsize (c)  }
\end{overpic}
\caption{\textbf{Training process.} \textbf{a.} Training loss across generations. The stepwise trend is due to the shifting training window. \textbf{b.} Performance of the i-th generation MCTS agent \rev{(continuous line) and number of games played by {\sc Olivaw} against itself during training in each generation (bar chart)}. The ELO ratings were computed using the evaluation games and the first match vs the national champion Alessandro Di Mattei. \textbf{c.} Absolute largest value drop in a game across generations. We show the averages using standard deviation as confidence interval.
}
\label{fig:trends}
\end{figure*}

\subsection{Low-cost faster training}

\label{sec-comparison}

With respect to AlphaGo Zero, {\sc Olivaw} introduces three main modifications in the training phase.  

As remarked, while the training set of AlphaGo consists only of
pairs of the kind $\{s, (\bm{\pi}_s, z_s)\}$, where $s$ is a move actually played during self-play and $z_s$ is the outcome at the end of the game, 
{\sc Olivaw} also considers pairs of the type $\{s, (\bm{\pi}_s, q_s)\}$, where $s$ is a position in the game tree that has been explored a number of times above a certain threshold.
The threshold value is set dynamically in order to have a training set twice the size of that used by AlphaGo. In other words, the number of pairs of type $\{s, (\bm{\pi}_s, q_s)\}$ is roughly equal to that of the pairs of type $\{s, (\bm{\pi}_s, z_s)\}$. The pairs added are the ones with the largest number of visits. 
Our approach was broadly inspired by the results reported in \citet{connect4MediumLessonsAlphaZero}.

Adding noisy pairs might not seem a good idea at first. In fact, 
using only the final outcome $z$ as a signal has a big drawback. In a game with multiple errors, every evaluation of an early position based on the final outcome is almost random, while $q$ offers a better assessment. On the other hand, $q$ suffers from the limited horizon of the \rev{search}; an early position with positive or negative consequences far ahead in the game may not be properly evaluated. So, a combination of the two signals might strike a better balance.

The second variation concerns a dynamic adjustment of the MCTS.
During the \emph{self-play phase} {\sc Olivaw} selects each move after 100, 200, or 400 MCTS simulations from the current game state, using a higher number of simulations in the higher generations\footnote{{\sc Olivaw} switched to 200 simulations between the 4th and the 5th generation, and to 400 simulations between the 11th and 12th generation.}, 
\rev{as opposed to AlphaGo Zero that stays with $1600$ simulations throughout the training run.}
The rationale is to move quickly away from early generations, where a dataset generated by low-depth MCTS still provides enough signal for an improvement, as noted by Wang, \emph{et al.} \citep{wang2019hyper}.

Finally, {\sc Olivaw} uses a dynamic training window. The training set is a sample of $16,384,000$ positions (minibatch size of 1024) from the games generated by the last generations. We gradually increase the generations included in the training window from the last two to the last five. The idea is to exclude quickly the games played by the first very weak generations. AlphaGo Zero uses as training set a sample of $2,048,000$ positions from the last $500,000$ games, always taken from games generated by the last 20 generations. This small modification proved effective in the Connect4 implementation of AlphaZero by \citet{connect4MediumLessonsAlphaZero}.

\section{Resources}

{\sc Olivaw} was entirely developed, trained, and tested on Colaboratory, a free Google cloud computing service for machine learning education and research \cite{colaboratory_carneiro2018performance}.

{\sc Olivaw} code is completely written in Python, from the core MCTS and Neural Network classes implemented in Numpy
and Keras, 
to the simple GUI based on Matplotlib. 
The \emph{self-play}, \emph{training} and \emph{evaluation phases} take place on three self-contained distinct notebooks sharing the same memory.
Concerning local resources, we took no more advantage than a laptop equipped with a browser and an Internet connection.

The hardware specifications of a Colaboratory virtual machine at the time of the training were:
\begin{itemize}
    \item CPU: 1 single core hyper threaded Xeon Processor, 2.3Ghz, 2 threads.
    \item RAM: $\sim$12.6 GB.
    \item Hardware accelerators (if used):
    \begin{itemize}
        \item GPU: 1 Nvidia Tesla K80, 2496 CUDA cores, 12GB GDDR5 VRAM.
        \item TPU v2-8: Google Tensor processing unit equipped with 8 TPU cores.  
    \end{itemize}
\end{itemize}
The generation of games during the \emph{self-play phase} is the most computationally expensive process of the learning algorithm.
In our hardware configuration, a single game takes from 6 to 30 seconds, depending on the number of MCTS simulations per move, 
due to the high number of GPU calls. \rev{The MCTS run on the CPU, while the deep network $f_\theta$ run on the GPU.}

Game generation can be naturally performed in parallel, by running several instances of the self-play notebook and saving the generated datasets on shared memory. This result has been achieved thanks to an informal crowd computing project, made possible by the ease with which Colaboratory can be shared. 19 people contributed to the generation of games with their Google accounts, also using smartphones.

The \emph{training phase} cannot be executed in parallel but can be accelerated using the TPU runtime of Colaboratory, the acceleration factor is approximately 10 with respect to a GPU K80. A single training phase required $\sim$1.5 hours.

The \emph{evaluation phase} consisted of 40 games between agents of different generations and took less than an hour of GPU run time.

\begin{figure*}[t]
\begin{overpic}[trim=0cm 0.9cm 3cm 0cm,clip,width=1.0\linewidth]{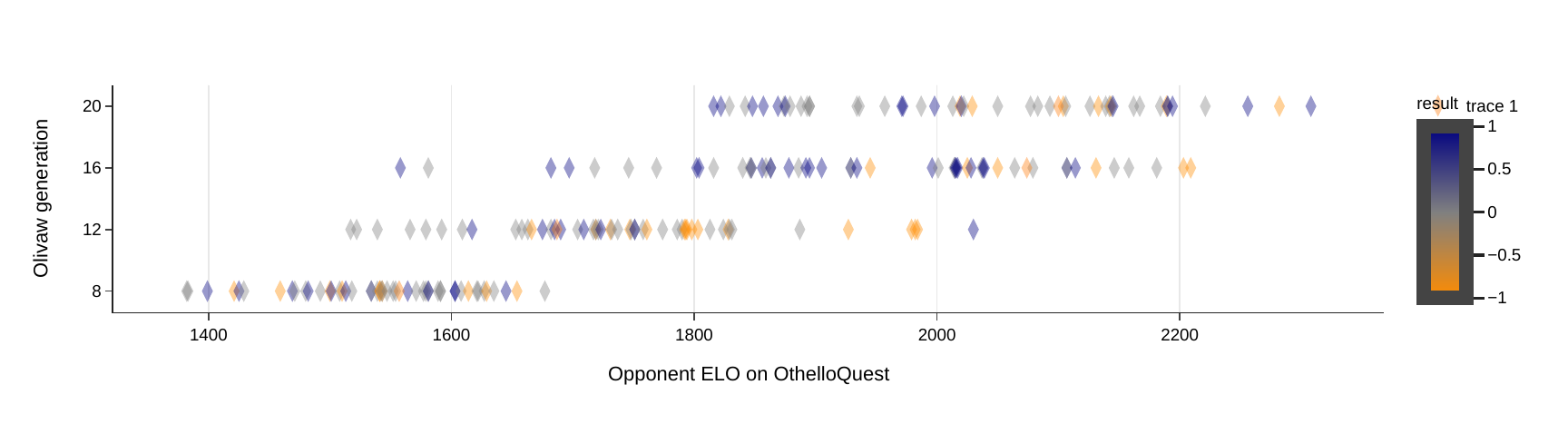}
\put(87,9){\color{white}
\frame{\includegraphics[trim=0cm 0cm 0cm 0cm,clip,width=0.1\linewidth]{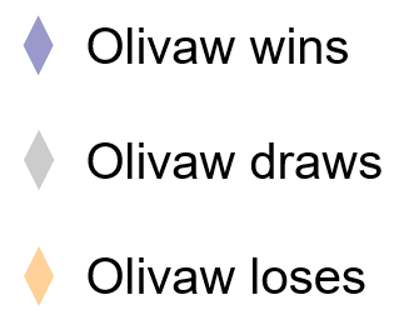}}}
\end{overpic}
\caption{\textbf{{\sc Olivaw}'s performance on OthelloQuest.} 
\rev{The score of different generations of {\sc Olivaw} on OthelloQuest. 
We report the outcome of the last 50 games played by every version of {\sc Olivaw} (the initial matches are warm-up games used by the platform to assess the strength of the player and thus are excluded).  Every agent played anonymously using 400 MCTS simulations per move.}}
\vspace{-0.4cm}
\label{fig_othelloquest}
\end{figure*}

\section{The training process}
\label{sec-training_process}

The version of {\sc Olivaw} discussed in this article is the result of a single training run lasting 30 days, 20 generations, and $\sim50,000$ games. We refer to the \emph{i-th generation} as the \emph{i-th successful update of the weights of $f_\theta$}.

Fine-tuning the hyperparameters for $8 \times 8$ Othello would have required a number runs incompatible with our main objective of mastering the game with limited resources.
Similarly, ablation studies to determine the effectiveness of our choices to improve the learning phase would have been prohibitively costly. As discussed however, these choices find good motivation in previous work, such as \citet{wang2019hyper} and \citet{connect4MediumLessonsAlphaZero}.

During the training phase, several interesting trends emerged (please refer to Figure~\ref{fig:trends}). Figure~\ref{fig:trends}~(a) plots the progress of the loss function across generations. Noticeable jumps take place when {\sc Olivaw} switches from one generation to the next. Recall that a training set consists of a batch of labeled games played by the most recent generations (the last two for the first generations and the last five later on). When the current oracle is defeated by a challenger, a new generation is born. The training set is updated by replacing the games played by the oldest generation with those of the most recent ones. The sudden drop in the loss can be ascribed to the improved quality of the new data set. This is an indication that {\sc Olivaw} is learning as generations go by. Other indications are given by the remaining two plots in Figure~\ref{fig:trends}. The plot in the middle simply reports the ELO rating of {\sc Olivaw} as generations go by \rev{and the number of games played by each generation against itself}. The rightmost plot shows the evolution of an interesting metric. Recall that the oracle provides two different types of evaluation given a state of the game: a probability distribution over the possible moves from that state and an assessment, ranging from $-1$ (certain defeat) to $+1$ (certain victory), of how good the current state is for {\sc Olivaw}. If the oracle is good, we expect the latter to change little from one move to the next. 
Conversely, a big drop from one state of play to the next is an indication that the oracle was unable to predict a very unfavorable move. The plot reports the maximum drop observed during a generation. It can be seen that the oracle is also improving according to this measure.

\begin{figure}[h]
  \begin{overpic}
		[trim=0cm 0cm 0.15cm 0cm,clip,width=0.99\linewidth]{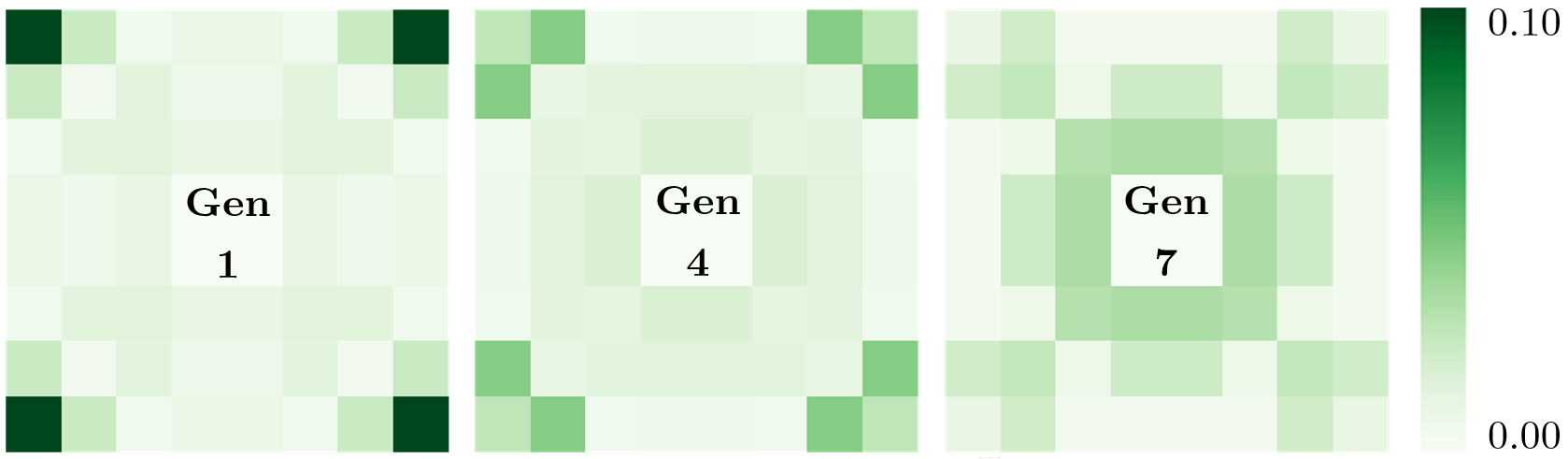}
		\end{overpic}
\caption{\textbf{Location of crucial moves by generation.} {\sc Olivaw} attributes high relevance to the conquest of the corners in the early generations, similarly to human beginners. In later generations, it shifts its ``attention'' towards the center of the board, as we would expect from a more experienced player.}
\label{fig:crucial}
\end{figure}

Another interesting fact is reported in Figure~\ref{fig:crucial}. It shows that, similarly to human beginners, early generations correctly but naively attach much significance to conquering the corners, gradually improving their appreciation of less conspicuous but more strategic positions in the middle of the board.

\subsection{Details to replicate training conditions:}
In the following, we report all salient hyperparameters to reproduce this work.

In each generation {\sc Olivaw} plays $\sim2500$ games against itself ($25,000$ in AlphaGo Zero), \rev{see Figure \ref{fig:trends}b}. As explained in Sec. \ref{sec-comparison}, each move is selected after 100, 200, or 400 MCTS simulations.
During the \emph{self-play phase}, the first 20 moves of each game are selected extracting at random according to $\bm{\pi}$ to favor exploration, the remaining moves are selected taking $\text{argmax}(\bm{\pi})$. As in AlphaGo Zero we use virtual losses \cite{segal2010scalability}.

When selecting a move from state $s$ with MCTS, in the root node we add Dirichlet noise to the prior probabilities computed by the neural network oracle:
\begin{equation}
    P(s_0, a) = (1 - \epsilon)p_a + \epsilon x_a \;\;\;\; \bm{x} \in \mathbb{R}^B
\end{equation}
Where $B$ is the number of legal moves from $s_0$, and
$\bm{x}$ is a point sampled from the symmetric Dirichlet probability distribution $X_\alpha$, with $\alpha=\mathrm{min}(1, 10 / B)$, we used an $\epsilon = 0.25$ as in AlphaGo Zero. 
A symmetric Dirichlet noise with an $\alpha < 1$ tends to unbalance the move probability distribution $\pi(a|s_0)$ towards a specific action, favoring an exploration in depth of the subsequent variant. In games with high branching factor, this behavior is desired to preserve the asymmetric nature of the MCTS search. So, the higher the branching factor of the game, the lower the $\alpha$ parameter of the symmetric Dirichlet noise used ($\alpha = 0.03$ in AlphaGo Zero).

When generating the dataset, not all games are played until the end to save computation. If during a game a player values a position under a resignation threshold $v_{\mathrm{resign}}$, the game ends and the opponent is considered the winner. $v_{\mathrm{resign}}$ is chosen automatically playing a fraction of the games until the end so that less than $5\%$ of those games could have been won if the player had not resigned. In early generations {\sc Olivaw} plays $10\%$ of the games until the end, \rev{i.e. the first 250 games. We increased} progressively this fraction to improve its strength in finals. This is because, differently from Go, Othello games are usually played until the very end, when no more moves are available.
(AlphaGo Zero plays always only $10\%$ of the games until the end).

In the \emph{training phase}, neural network parameters $\theta$ are optimized using stochastic gradient descent with momentum $\lambda=0.9$ and a stepwise learning rate annealing. Starting from a learning rate of 0.003, {\sc Olivaw} switched to 0.001 in the 4th generation and to 0.0001 in the 11th generation. 

Concerning the architecture, {\sc Olivaw} uses a Residual Network \cite{He2015DeepRecognition} as AlphaGo Zero. The game state input $s$ is a $8 \times 8 \times 2$ binary tensor in {\sc Olivaw}, in contrast to the deeper $19\times 19 \times 17$ binary tensor of AlphaGo Zero. We do not need layers representing past positions in Othello since the game state is fully observable from the board position. Even the information on the player's turn is not necessary since Othello is symmetrical with respect to the player, we can assume that the turn player is always white, flipping all the discs if it is the turn of black.

Therefore the input $s$ is processed by a single convolutional block and then by a residual tower of 10 residual blocks (39 in the strongest version of AlphaGo Zero). The output of the residual tower is then processed by the value head and the policy head that output respectively the value $v$ of the position and the move probabilities $\bm{p}$. The structure of each block coincides with the correspondent one of AlphaGo Zero, except for the output shape of the policy head, $8^2 + 1=65$ in {\sc Olivaw} instead of $19^2 + 1 = 362$.

\begin{figure}[t]
\vspace{-0.1cm}
\includegraphics[trim=0cm -1.8cm 0cm 0cm,clip,width=0.21\textwidth]{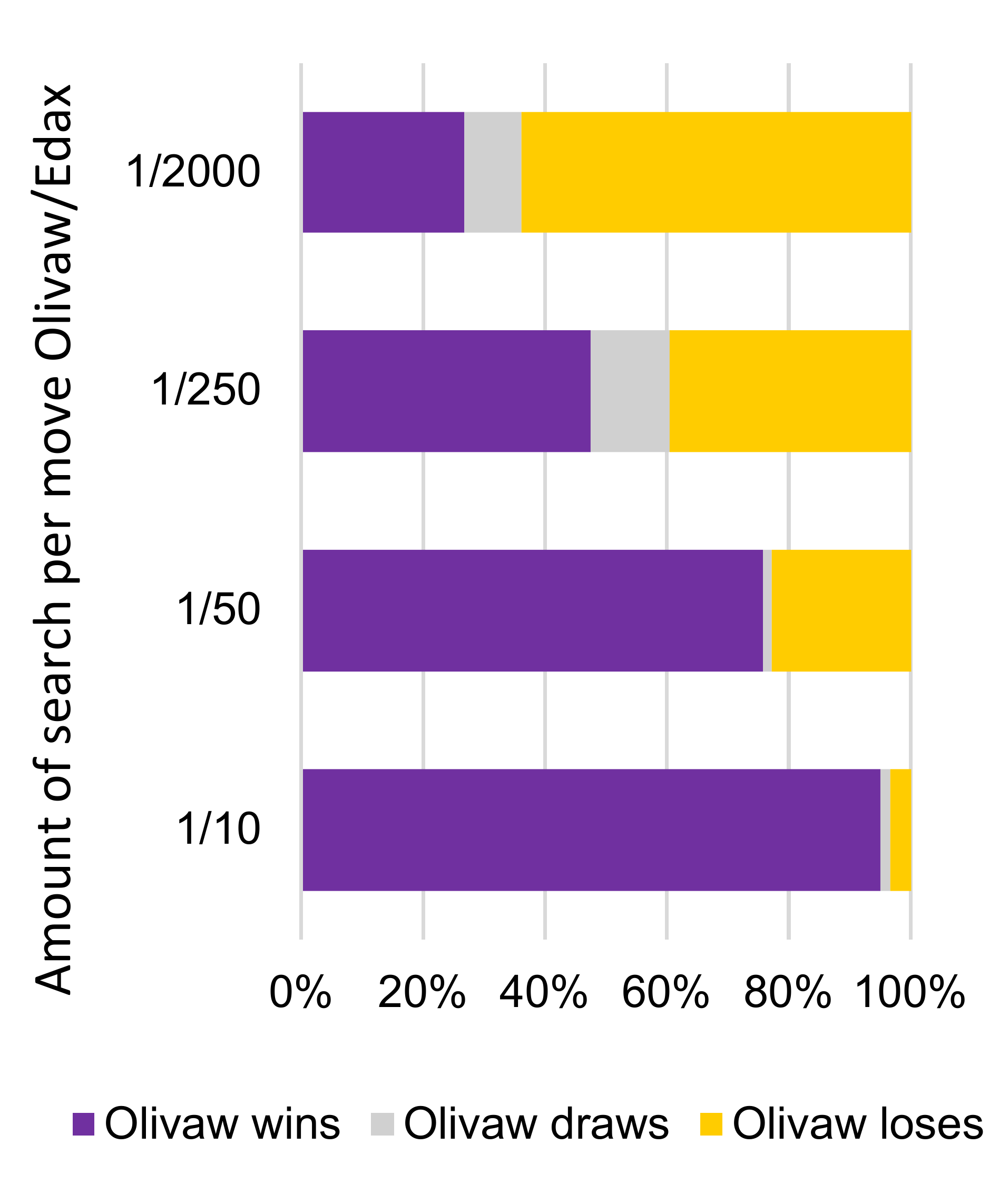}
\includegraphics[width=0.27\textwidth]{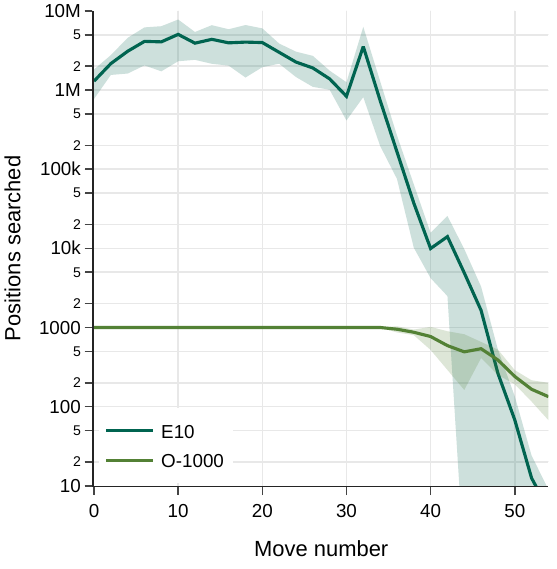}
\hspace{-0.6cm}
\centering
\vspace{-0.3cm}
\begin{overpic}
		[trim=10cm 34cm 2cm 0cm,clip,width=1.0\linewidth]{images/crucial_moves.png}
		\put(22,2){\scriptsize (a)  }
		\put(75,2){\scriptsize (b)  }
\end{overpic}
\caption{\rev{\textbf{{\bfseries\scshape Olivaw} vs Edax. a.} 
Outcome of four $100$-game series between O-$1000$ and E$4$, E$6$, E$8$, and E$10$. 
\textbf{b.} Number of game-tree positions  searched by E$10$ and O-$1000$ for each move, averaged over 10 games.}}
\label{fig_edax}
\end{figure}

\section{Attaining world-class level in Othello}
\rev{
We tested the strength of {\sc Olivaw} in three different ways. First, 
by playing anonymous games on the web platform OthelloQuest. Second,  by pitting it against Edax, one of the strongest open-source Othello engine.  And finally, with two matches against top-notch human players: a national champion and a former world champion.}
\rev{
\subsection{Matches on the web platform OthelloQuest}
}
\rev{During training, the strength of {\sc Olivaw} was tested with a series of anonymous online games against human players on OthelloQuest, a popular Othello platform which is also widely used by top human players. {\sc Olivaw} was deployed at generation 8,12,16, and 20 with the number of explorable nodes in the game tree set at 400. The duration of every game was set to 5 minutes. In Figure \ref{fig_othelloquest} we report the score of the last 50 games played by each version of {\sc Olivaw}. We excluded the positioning games used by the platform to assess the level of a new player, which is why we do not report the games between {\sc Olivaw} 20 and opponents ranked at 1400. A clear improvement can be observed as generations progress. Performance ratings
\footnote{This rating is estimated from the games of a single event only, see \url{https://en.wikipedia.org/wiki/Elo_rating_system}.}
after these 50 games of generations 8, 12, 16, and 20 are, respectively, 1557, 1687, 2074, and 2100.}
\rev{
\subsection{Matches against Edax}}
We tested {\sc Olivaw} against Edax \cite{Edax}, arguably the strongest open-source Othello engine \cite{liskowski2018learning}. Like other top traditional engines, Edax is based on a highly optimized alpha-beta tree search (negamax) \cite{knuth1975analysis} using tabular value functions\footnote{Unfortunately, a detailed description of how these value functions are obtained is not publicly
available, \rev{see \cite[Section 2]{liskowski2018learning} for further details.}}. 
\rev{In what follows E$k$ denotes the version of Edax in which the depth of the alpha-beta search in the game tree is limited to $k$. In our comparison, E$4$, E$6$, E$8$ and E$10$ were used.}
For the games we report, Edax used no opening book. Edax is a deterministic engine and when it played against {\sc Olivaw} the same single game was repeated again and again. To circumvent this problem we switched to random XOT openings, a popular Othello variation where the first 8 moves are chosen at random from a list of 10,784 sequences ending in an almost even position, i.e. positions judged between -2 and +2 discs advantage for black by Edax at search depth 16. 

\rev{The four versions of Edax were pitted against four versions of the 20th generation of {\sc Olivaw}. The four versions deployed correspond to the maximum number of nodes of the game tree allowed to be explored and that were set to $400, 1000, 2500,$ and $10$ thousand. The resulting versions of {\sc Olivaw} are referred to as O-$400$, O-$1000$, O-$2500$, and O-$10$T.}

\rev{Several aspects must be considered in order to set up a comparison between Edax and {\sc Olivaw} that is informative as well as feasible. As far as the former aspect is concerned, setting limits in terms of wall-clock time for the two agents would not be very informative. They use different hardware (GPUs vs CPUs) and are written in different programming languages. This is why we opted for a machine-independent measure of computational effort: the number of explored nodes in the game tree to decide the next move. (In the case of Edax the depth of the alpha-beta search translates into number of nodes explored, see below). 
As for the latter, it seems reasonable to assume that, while increasing the budget of explorable nodes makes an agent stronger, sooner or later a plateau must be reached whereby increasing the budget does not translate in more strength. In order to compare the best of the two agents therefore, it is tempting to determine such a plateau and have the two resulting top versions play each other. This approach however is completely unfeasible in terms of resources, especially in our constrained framework.}

\begin{figure}[t]
\includegraphics[width=0.49\textwidth]{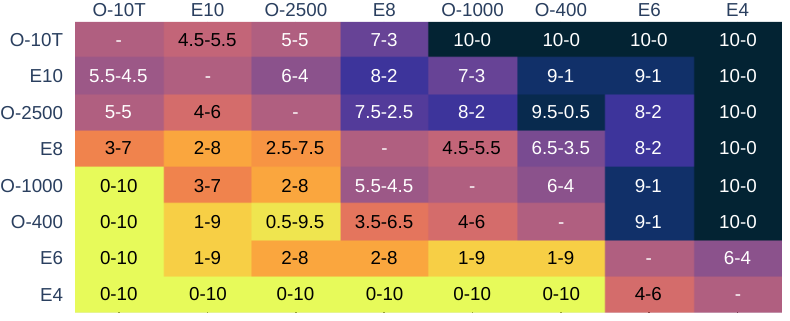}
\centering
\caption{\rev{\textbf{{\bfseries\scshape Olivaw} vs Edax: The Tournament.}  
Outcome of a tournament among the {\sc Olivaw}-type agents (O-$400$, O-$1000$, O-$2500$, and O-$10$T) and the Edax-type agents (E$4$, E$6$, E$8$, and E$10$), whereby every agent plays a $10$-game series against every other agent. Darker (resp. lighter) colours indicate a favourable (resp. unfavourable) outcome for the agents listed on the left column.}
}
\label{fig-tournament}
\end{figure}

\rev{Thus, we settled for the following: a tournament in which the eight agents played against each other. A match between two agents consists of a $10$-game series. For every game, we assigned 1 point for a victory, $0.5$ points for a draw, and $0$ points for a defeat.
A widespread assessment among accomplished Othello players is that these four versions of Edax have a strength that gradually increases from good (E$4$) to exceptionally strong (E$10$) (to the level of the best human players, if not stronger), thus providing a good benchmark to assess {\sc Olivaw}'s strength. The outcome of the tournament is reported in Figure~\ref{fig-tournament} (matches) and Table~\ref{tab-tournament} (leaderboard).}

\rev{Conforming to intuition,  it is apparent that as the exploration budget increases all agents become stronger.} 
\rev{The main take-away point however is that 
{\sc Olivaw} is competitive against Edax in spite of the fact that it explores far fewer nodes in the game tree than its opponent. On average, E$4$, E$6$, E$8$ and E$10$ explored $10,000$, $50,000$, $250,000$, and $2,000,000$ nodes to make a move, respectively, whereas we recall that the four versions of {\sc Olivaw} explore at most $400, 1000, 2500,$ and $10,000$ nodes per move.
This overall conclusion is further exemplified by Figure~\ref{fig_edax}a, which shows the outcome of four $100$-game series between O-$1000$ versus E$4$, E$6$, E$8$ and E$10$.
}

\rev{As Table~\ref{tab-tournament} shows, O-$10$T won the tournament, even if it lost the series against E$10$ by a narrow margin. This is because, interestingly, weak versions of {\sc Olivaw} still managed to occasionally beat the strongest version of Edax.}

\rev{It is perhaps worth noting that, unlike {\sc Olivaw}, Edax can take advantage of heuristics fine-tuned by humans to choose when to expend more search. Figure \ref{fig_edax}b shows the average number of positions searched  during the series between  E$10$ and O-$1000$.  Besides the huge difference in game-tree exploration effort, notice the peak 20 moves before the end. It is when Edax is programmed to search deeper in preparation for the endgame.}

\rev{Overall, the results reported in this section indicate that {\sc Olivaw} is a very strong player, possibly at the level of top human players. We test this hypothesis in the next section.}
\begin{table}[]
\centering
\caption{\rev{Leaderboard of the tournament between Olivaw and Edax}}
\label{tab-tournament}
\begin{tabular}{ccc}
\hline
\textbf{Points} & \textbf{Player} & \textbf{$\max($positions searched per move$)$} \\ \hline
56,5            & O-10T      & $1.0 \times10^4$                                                                     \\
54,5            & E10         & $(8.3 \pm 3.2) \times 10^6$                                                          \\
52              & O-2500     & $2.5 \times 10^3$                                                                    \\
36,5            & E8 & $(1.1 \pm 0.3) \times 10^6$                                                          \\
35,5            & O-1000     & $1.0 \times 10^3$                                                                    \\
28              & O-400      & $4.0 \times 10^2$                                                                    \\
13              & E6        & $(1.0 \pm 0.2) \times 10^5$                                                          \\
4               & E4        & $(1.7 \pm 0.2) \times 10^4$                                                          \\ \hline
\end{tabular}
\end{table}
\rev{
\subsection{Matches against top-notch human players}
}
As a final, and much more enjoyable, battery of tests, we organized three live series against top human players with the support of the Italian Othello Federation. Two were against the 2019 Italian champion Alessandro Di Mattei, ranked among the top 150 Othello players in the world, and one, more formal challenge, against the former World champion Michele Borassi, ranked in the top 50 and who finished in the top five in his last World Othello championship appearance in 2018\footnote{World Othello ratings at the time of matches.}.

\textbf{Matches against a national champion.}
Two informal best-of-five series were organized against the Italian champion Alessandro Di Mattei. They took place between 27 November and 4 December 2018 in Rome.
The first series against Di Mattei saw the 14th generation of {\sc Olivaw} very close to the national champion, who emerged victorious after two initial draws. A post-match analysis of the games showed {\sc Olivaw} losing from winning positions in the final moves. This feedback led to the decision of simulating all the subsequent \emph{self-play games} until the very end, to strengthen {\sc Olivaw}'s late game. 
The second series against generation 18 ended with a resounding 4-0 victory for {\sc Olivaw}. This boosted our confidence and we threw down the gauntlet against a former world champion. Table \ref{tab-games} shows the matches against Di Mattei.

\textbf{Challenging a former World champion.}
Former world champion Michele Borassi picked up the gauntlet and a formal series was organized. The match was sponsored by the Sapienza Computer Science department and was open to the public \rev{and streamed live over the internet}. 
The formula was a best-of-three with 30 minutes for each player for the whole game, as in the world championship. After the good results against Di Mattei, we decided to keep the MCTS simulations of the game tree at 1000 -\rev{a very small amount of search per move, insufficient to defeat even beginners for traditional programs like Edax}- and to stop the training at generation 20, after $\sim50,000$ games simulated in \emph{self-play}. 
\rev{In short, Borassi played against O-$1000$}.

{\sc Olivaw} won the first game as black, losing the second and third one with white. The final score was thus 2-1 against {\sc Olivaw}. All matches are shown in Table \ref{tab-games}. 

\begin{figure}[h]
\includegraphics[trim=1.9cm 0.5cm 0cm 0.cm,clip,width=0.47\textwidth]{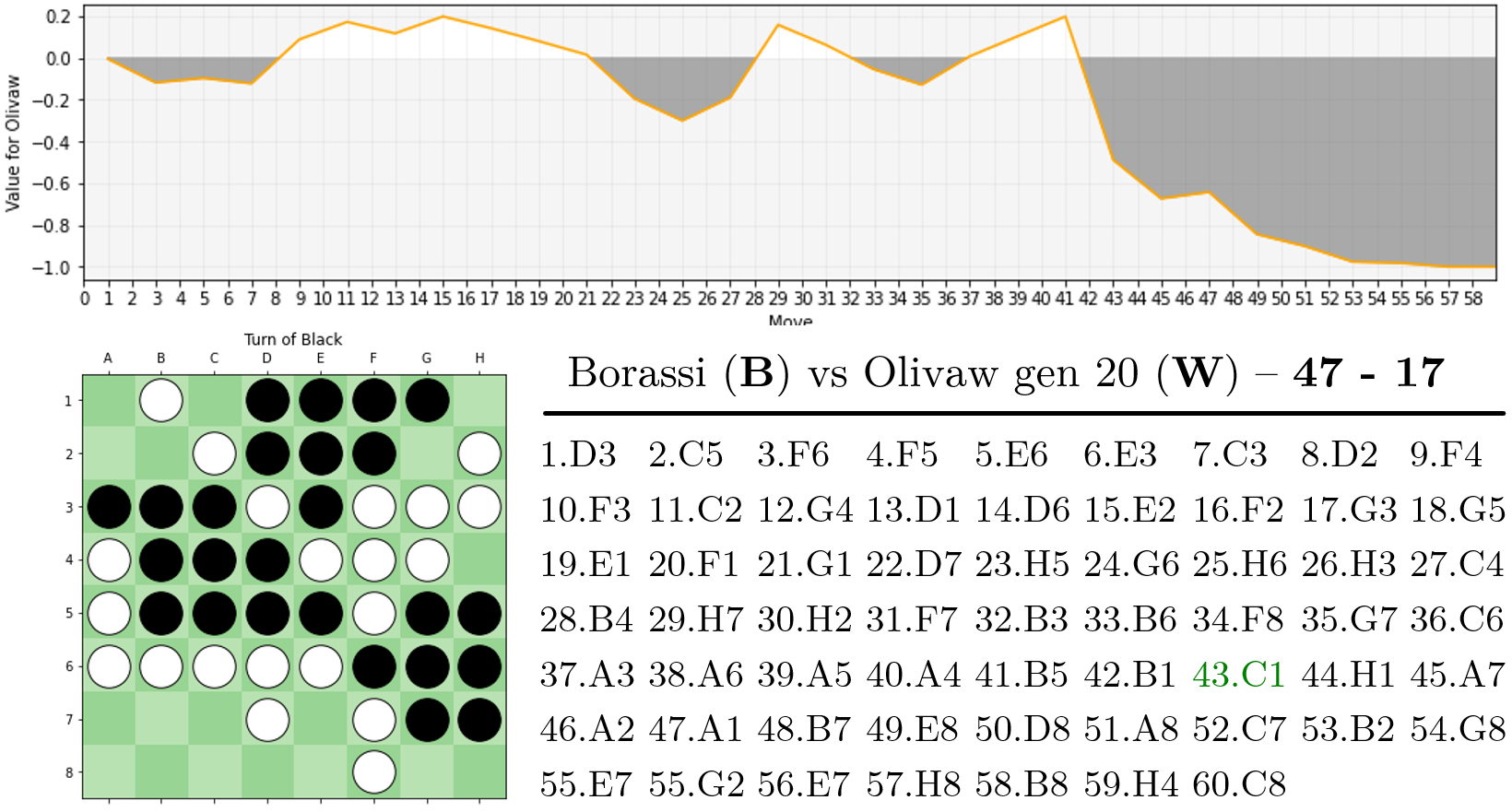}
\centering
\caption{\textbf{Match between {\sc Olivaw} generation 20 and the former World champion Michele Borassi, final game.} The graph shows the confidence of winning of {\sc Olivaw}; -1 a sure defeat, +1 a sure victory.}
\label{fig-Borassi}
\vspace{-0.3cm}
\end{figure}

Othello players may find interesting move 43. C1 of Borassi in game~3. It is a highly counter-intuitive move and the only option to avoid defeat: with a masterstroke the former world champion snatched victory from the jaws of defeat. Borassi spent more than one-third of its time on that single move. {\sc Olivaw} did not see it coming, as evidenced by the large value drop recorded on move~43 (see Figure \ref{fig-Borassi}). 
An act of \rev{digital hubris} that proved fatal.

\begin{table*}[t]

\caption{ {\sc Olivaw} games versus top Othello players.}

\setlength{\tabcolsep}{6pt}
\centering
\resizebox{\linewidth}{!}{\begin{tabular}{ll|c|l}
\multicolumn{4}{c}{First match against the national champion Alessandro Di Mattei - Best of 5 - 2018/11/27} \\
\hline
Di Mattei & {\sc Olivaw} 14 & 32-32 & {\tiny C4E3F6E6F5C5C3C6D3D2E2B3B4C2B6A4B5D6A3A5A6F3F4G4F7D1F1D7E1C1B1G6C7E7F8D8H6F2G1G5C8B8G7B7E8G2A8A7H1G3H2H3H4B2A2A1PAH5PAH8G8H7 \par}  \\ 
{\sc Olivaw} 14 & Di Mattei & 32-32 & {\tiny D3C3C4C5B4D2D6C6E6D7B5A3C7C8B6A6E3E7A4F2F8F5F6F4A2E8F7G5G6H6C2B7F3A5A8A1E2B3D8B8B2G4G3B1C1H3F1E1D1G7H4H5A7G1H8G8H7G2H2PAH1 \par}  \\ 
Di Mattei & {\sc Olivaw} 14 & 43-21 & {\tiny C4E3F6E6F5C5C3C6D3D2E2B3C1C2B4A3A5B5A6F4F3B6E7D1E1G5G4G6H5D6B1H3H4H6F7F8D8A4A2E8G8G3D7C8B8G7H8H7A7C7H2G2G1B7A8F1F2H1PAA1PAB2 \par}  \\ 
{\sc Olivaw} 14 & Di Mattei & 36-28 & {\tiny F5F6E6D6E7G5C5C6C4F3D7D8C7C8F4B6G6G4H5H6H4H3G3B5E3B3B4C3A5A6D3C2D2D1A3H2C1B1F7A4A7B2E2E8G8G7F8H8H7G2F2B7A8B8H1F1G1E1A1A2 \par}  \\ 
Di Mattei & {\sc Olivaw} 14 & 33-31 & {\tiny C4E3F6E6F5C5F4G6F7C3H6G4G3D7E7F3F2H3D3E2E1C6D6G5D2C7C8C2B1E8B8F1G1G8H5H4B7B5A5B4F8B6D8A8H2C1D1H7H8G7B2B3A3A4A7A6PAA2PAG2H1PAA1 \par}  \\ 
\hline
\hline
\noalign{\vskip 1mm}    
\multicolumn{4}{c}{Second match against the national champion Alessandro Di Mattei - Best of 7 - 2018/12/04} \\
\hline
Di Mattei & {\sc Olivaw} 18 & 27-37 & {\tiny C4E3F6E6F5C5C3C6D3D2E2B3B4A3E7C2D6F1D1F4E1C1B6F3B2D7C8G5H4C7D8G6H6H5G4H3H2B7B1B5A8A1A5E8F7G7A4F8H7A6A2H8G8H1G3F2A7B8G1G2 \par}  \\ 
{\sc Olivaw} 18 & Di Mattei & 40-24 & {\tiny E6F6F5D6E7G5C5C6E3C4D7F8B4D3C3A3B5B3B6C8A4A5A6A7F4C7G6H6F7G8H4H5H7C2D2F2F3D1E2G2E1C1B1G4F1B2A1A2A8G3E8D8B8B7H8G7H1H2H3G1 \par}  \\ 
Di Mattei & {\sc Olivaw} 18 & 27-37 & {\tiny C4E3F6E6F5C5C3C6D3D2E2B3C1C2B4A3A5B5A6B6A4A7E7E1D6D7C8F3C7F8F7G5H4G6H5D1F1F2B1H6H7G4H3F4G3E8B7D8G8B2B8G7A2A1PAG1G2H2H1 \par}  \\ 
{\sc Olivaw} 18 & Di Mattei & 45-19 & {\tiny C4E3F6E6F5C5C3B4D3C2D6F4E2F3D2C6G5G4F2G3H3H4B5H6A3B6D7A4A5A6B3C1E1A2D1F1G2E7E8F7H2C8F8D8C7G8B7G6H5H1G1B8G7A8A7B2A1B1PAH8H7 \par}  \\ 
\hline
\hline 
\noalign{\vskip 1mm}    
\multicolumn{4}{c}{Match against the former World champion Michele Borassi - Best of 3 - 2019/01/19} \\
\hline
{\sc Olivaw} 20 & Borassi & 35-29 & {\tiny E6F4C3C4D3D6F6E7F3C5F5G4B5G5C6E3D7B4G6F7E2C8B6C7A5H5E8F2B3F8D8E1H6A4A3C2C1A7H3H4A6A2D2D1G3G2H1H2B2B8A8B7A1B1G8G7H8H7F1G1 \par}  \\ 
Borassi & {\sc Olivaw} 20 & 35-29 & {\tiny D3C5F6F5E6E3C3D2F4F3C2G4D1D6E2F2G3H4G1F7H3G6C4H2G5B4H5H6E7B3C7C6D7D8B5A5E8F8A6C8A4F1E1B1B6A2C1H1A3A7G7G8H8H7B7B2A1A8B8G2 \par}  \\ 
Borassi & {\sc Olivaw} 20 & 47-17 & {\tiny D3C5F6F5E6E3C3D2F4F3C2G4D1D6E2F2G3G5E1F1G1D7H5G6H6H3C4B4H7H2F7B3B6F8G7C6A3A6A5A4B5B1C1H1A7A2A1B7E8D8A8C7B2G8G2E7H8B8H4C8 \par}  \\ 
\hline
\hline 
\end{tabular}
}

\label{tab-games}
\end{table*}

\section{Conclusion}
\rev{After one month of training, using only free, and quite limited, cloud computing resources, {\sc Olivaw} achieved world-class level in the game of Othello.  The high ELO rating reached on the popular web platform OthelloQuest, the winning challenges against the strongest open-source Othello engine Edax, the victory against national champion Alessandro Di Mattei, and the honorable defeat against former World champion Michele Borassi corroborate this assessment.}

\rev{Differently from traditional Othello engines like Edax, {\sc Olivaw} reached these results using a minimal amount of search per move,
and learned its value function completely from scratch, like AlphaGo Zero.
The amount of training required was $\sim50,000$ games played against itself. Differently from its illustrious predecessor, {\sc Olivaw} maximizes the information extracted from each game by adding to the training set also positions not played but largely explored by the agent.}

The level of play reached by {\sc Olivaw} is not yet superhuman however. This would be the natural next step for {\sc Olivaw} and is left for future work. To be interesting, such an achievement should come within the same resource-limited paradigm, using a limited amount of computational time and power in the training phase and a small number of MCTS simulations per move during  matches against a human opponent.

\section*{Acknowledgments}

This research was partially supported by a Google Focused Award and by BICI, the Bertinoro International Center for 
Informatics.

We thank Alessandro Di Mattei and  Michele Borassi for agreeing to play against {\sc Olivaw}; Paolo Scognamiglio and Alessandro Di Mattei for introducing us to the Othello world and the helpful discussions on {\sc Olivaw}'s style of play, Roberto Sperandio for the gripping match commentary, Benedetto Romano for the discussion on traditional engines, Leonardo Caviola for having introduced the public to Othello during the match-day, and the whole Italian Othello Federation.

We thank all the game generators:
Dario Abbondanza,
Armando Angrisani,
Federico Busi,
Silva Damiani,
Paola D'Amico,
Maurizio Dell'Oso,
Federico Fusco,
Anna Lauria,
Riccardo Massa,
Andrea Merlina,
Marco Mirabelli,
Angela Norelli,
Oscar Norelli,
Alessandro Pace,
Anna Parisi,
Tancredi Massimo Pentimalli,
Andrea Santilli,
Alfredo Sciortino, and
Tommaso Subioli.

\printbibliography

\end{document}